\documentclass{article}

\usepackage[final]{tackling_climate_workshop_style}

\usepackage[utf8]{inputenc} 
\usepackage[T1]{fontenc}    
\usepackage{hyperref}       
\usepackage{url}            
\usepackage{booktabs}       
\usepackage{amsfonts}       
\usepackage{nicefrac}       
\usepackage{microtype}      
\usepackage{amsmath}
\usepackage{bbm}
\usepackage{graphicx}
\usepackage{lipsum}
\usepackage{caption}
\usepackage{subcaption}
\usepackage{xcolor}
\usepackage{booktabs}
\usepackage{mathtools}

\title{Probability calibration for precipitation nowcasting}

\author{%
  Lauri Kurki \\
  Vaisala\\
  Espoo, Finland \\
  \texttt{lauri.kurki@vaisala.com} \\
  \And
  Yaniel Cabrera \\
  Vaisala \\
  Espoo, Finland \\
  \texttt{yaniel.cabrera@vaisala.com} \\
  \And
  Samu Karanko \\
  Vaisala \\
  Espoo, Finland \\
  \texttt{samu.karanko@vaisala.com}
}

\begin{document}

\maketitle

\begin{abstract}
    Reliable precipitation nowcasting is critical for weather-sensitive decision-making, yet neural weather models (NWMs) can produce poorly calibrated probabilistic forecasts. Standard calibration metrics such as the expected calibration error (ECE) fail to capture miscalibration across precipitation thresholds. We introduce the expected thresholded calibration error (ETCE), a new metric that better captures miscalibration in ordered classes like precipitation amounts. We extend post-processing techniques from computer vision to the forecasting domain. Our results show that selective scaling with lead time conditioning reduces model miscalibration without reducing the forecast quality.
\end{abstract}

\section{Introduction}

Precipitation nowcasting—forecasting the immediate future with lead times of up to four hours at high temporal and spatial resolution— supports time-sensitive decisions in disaster response, transportation safety, urban drainage management, and winter road maintenance \cite{gen-models, nowcast-challenge, nowcastnet}. As climate change drives shifts in precipitation patterns and increases the frequency of extreme weather events having skillful nowcasts is ever more important.  

Recent years have seen rapid advances in precipitation forecasting through deep neural networks (DNNs) \cite{shi2025deeplearningfoundationmodels, aabonev2023modelling, mcnally2024datadrivenweatherforecasts, bi2023accurate}. Neural weather models (NWMs) are state-of-the-art systems and are being deployed operationally by both industry and meteorological agencies \cite{andrychowicz2023deeplearningdayforecasts, lang2024aifsecmwfsdatadriven, adrian2025data}. Many applications demand not only accurate but also probabilistic forecasts, where predicted probabilities reflect the true likelihood of events. Two main approaches exist: generating ensembles of perturbed predictions, or directly building probabilistic models \cite{bi2023accurate,  andrychowicz2023deeplearningdayforecasts, price2025probabilistic}. In this work, we focus on the latter.

A key requirement for reliable probabilistic forecasts is calibration—the alignment between predicted probabilities and observed event frequencies. For a model that predicts a class $\hat Y$ with probability $\hat P$, perfect calibration is formally defined as $\mathbb{P}( \hat Y = Y  | \hat P = p) = p,\ \text{for all}\ p\in[0,1]$ and all class labels $Y \in \left\{0, \dots, K -1 \right\}$ \cite{Nixon_2019_CVPR_Workshops}. For classification and semantic segmentation models, calibration is often assessed via the \emph{expected calibration error} (ECE)
\begin{align}
    \operatorname{ECE} = \sum_{b=1}^B\frac{n_b}{N}\left| \operatorname{acc}(b) - \operatorname{conf}(b)\right|,
\end{align}
where $n_b$ is the number of predictions in bin $b$, $N$ is the total number of data points, and $\operatorname{acc}(b)$ and $ \operatorname{conf}(b)$ are the mean accuracy and confidence in that bin \cite{Nixon_2019_CVPR_Workshops, guo, wang10205155, NEURIPS2019_f8c0c968}. However, ECE has well-known shortcomings, especially in multiclass problems with ordered categories, such as precipitation amounts \cite{Nixon_2019_CVPR_Workshops, gawlikowski2023survey}. In this context, the metric’s focus on the predicted class and its associated confidence obscures miscalibration across the full range of precipitation bins. For example, a winter maintenance operator must understand probabilities across multiple thresholds (e.g., 1 mm versus 10 mm of snowfall), not just the most likely category. Static calibration error (SCE) \cite{Nixon_2019_CVPR_Workshops} extends ECE to a multiclass setting but it's still intended only for independent classes. Therefore, for a model predicting a probability vector $\hat P(r)$ over precipitation rates $r\in[R_1,\dots,R_K]$, the calibration is better defined by
\begin{align} \label{eq:new-error}
    \mathbb{P}( r > R \ | \ \hat P (r > R) = p) = p,\ \text{for all}\ p\in[0,1],\ R\in[R_1,\dots,R_K].
\end{align}
In other words, given 100 predictions for precipitation $r > R$ each at confidence 0.8, we expect that for 80 \% of those predictions, precipitation will exceed $R$.

In this note, we address the miscalibration of probabilistic NWMs in precipitation nowcasting. We introduce the \emph{expected thresholded calibration error} (ETCE) as a more appropriate metric for assessing miscalibration in our setting. We study post-processing techniques designed to adjust model confidence values to better match observed precipitation frequencies. While there is active research on the topic, existing calibration methods are limited—particularly in computer vision—and, to our knowledge, absent in the context of NWMs \cite{wang10205155, 9710154, ding_local}.

\section{Methodology}
\subsection{Expected thresholded calibration error}

To estimate the calibration error of Eq. \ref{eq:new-error}, we first compute confidences for the thresholded prediction $\hat P(r > R_k)$ for all $k\in [1, \dots, K]$. Then, for each threshold we bin the predictions by predicted confidence into $B$ evenly spaced bins. Finally, for each bin $b$ and threshold $R_k$ we compute the mean predicted confidence $\operatorname{conf}(b,R_k)$ and mean accuracy $\operatorname{acc}(b,R_k)$. Then, we compute the difference between predicted confidence and accuracy as
\begin{align}
    \operatorname{ETCE} = \frac{1}{K}\sum_{k=1}^K \sum_{b=1}^B w_b\left|\operatorname{acc}(b, R_k) - \operatorname{conf}(b,R_k)\right|.
\end{align}
where $w_b$ are bin weights. We expand the confidence and accuracy terms in the appendix (see Section \ref{sec:details}). In this application, we apply uniform weighting $w_b = 1 / B$ as precipitation is a rare event and weighting by the number of samples would emphasize dry events in the metric. In our experiments we have enough data to select $B = 20$ (See Figure \ref{fig:counts} in the Appendix).

\begin{figure}
    \centering
    \includegraphics[width=0.7\linewidth]{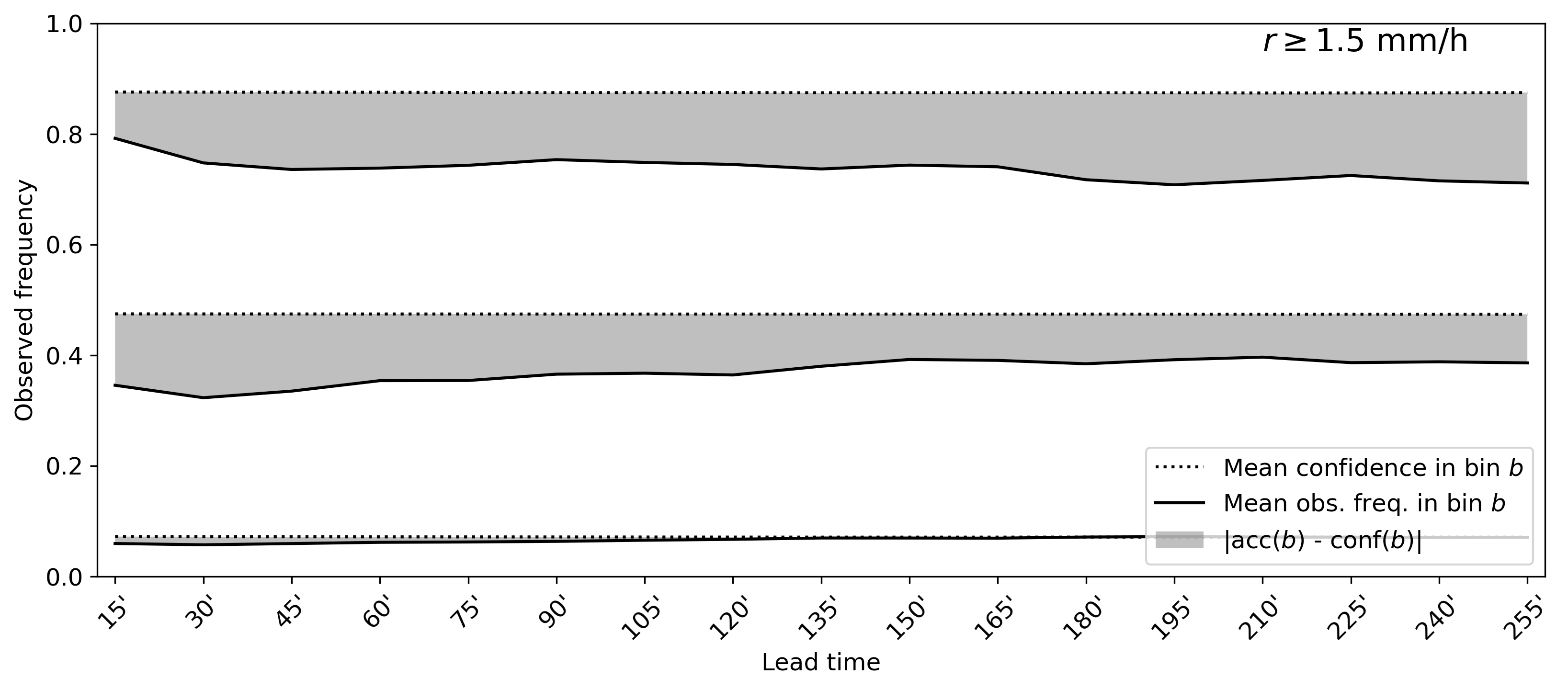}
    \caption{Miscalibration diagram at precipitation threshold $1.5 \operatorname{mm/h}$ and selected confidence bins $[0.05, 0.10],\ [0.45,0.50],\ [0.85,0.90]$. The mean confidence and mean observed frequency for each bin are depicted by dashed and solid curves respectively.}
    \label{fig:metric}
\end{figure}

To build intuition on ETCE, we illustrate the miscalibration between predicted confidence and observed frequency at fixed threshold $R_k = 1.5$ mm/h for three selected bins in Figure \ref{fig:metric}. The calibration error $|\operatorname{acc}(b)-\operatorname{conf}(b)|$ corresponds to the filled area between observed frequency and confidence. In this example, the observed frequency is smaller than the average confidence in the same bin meaning that the model is overconfident. For a better calibrated model, the area between observed frequency and predicted confidence is smaller.

\subsection{Calibration methods}

In the literature there are multiple post-processing tools for calibrating probabilistic models in deep learning. Here we describe the calibration methods we extended and tested in the forecasting domain.

\textbf{Temperature scaling} (TS), and its variations, is calibration method which has been shown to be effective for classification tasks \cite{guo, hinton2015distillingknowledgeneuralnetwork, balanya2024adaptive}. In TS, a single parameter $T\in\mathbb{R}^+$ is learned---typically by minimizing the negative log-likelihood---to scale the predicted probability $\boldsymbol{\hat p} = \sigma_{\operatorname{softmax}}(\boldsymbol{z} / T)$. In segmentation tasks, one temperature is optimized to scale predicted probabilities of pixels and all samples.

\textbf{Local temperature scaling} (LTS) \cite{ding_local} is an extension of TS in which a different temperature is applied for each pixel $x$ in a sample. In this approach, we need to learn a regressor for mapping a logit vector $\boldsymbol{z}$ to a temperature value $T$. LTS is proposed to better calibrate segmentation models where especially the label boundaries are often miscalibrated. In this work and in \cite{ding_local}, LTS learns the temperature mapping using a small hierarchical CNN. Different from \cite{ding_local}, we only use the predicted logits as input and also condition the regressor with lead time using Feature-wise Linear Modulation (FiLM) \cite{perez2017filmvisualreasoninggeneral} which applies affine transformations to intermediate feature maps based on external information.

\textbf{Selective scaling} (SS) \cite{wang10205155} is based on the observation that the major cause of neural network miscalibration is overconfidence on mispredictions. SS uses a classifier on the logits $\boldsymbol{z}$ to first detect mispredictions of the base model, and then applies scaling with temperature $T > 1$ only on the mispredictions to reduce overconfidence to obtain the calibrated probability vector $\boldsymbol{ \hat {p}}$,
\begin{align}
    \boldsymbol{ \hat {p}} = \begin{cases}
        \sigma_{\operatorname{softmax}}(\boldsymbol{z}),\ & \text{if}\ \hat y = y \\
        \sigma_{\operatorname{softmax}}(\boldsymbol{z} / T),\ & \text{if}\ \hat y \neq y. \\
    \end{cases}
\end{align}
In \cite{wang10205155} the classifier is a 3-layer MLP conditioned on the logits. We augment that architecture by using FiLM to pass the lead time encoding. We also investigate enhancing the classifier's spatial view through larger attention-based architectures.

\subsection{Data and model}
{\bf Base model.} In our experiments we hold the probabilistic base model fixed. The model is conditioned on lead time so the predictions are independent across lead times. The architecture has three main components: a spatial encoder consisting of a sequence of convolutional layers to downsample the input data from $512 \times 512$ pixels to $64 \times 64$ with 512 channels; an attention block composed of four axial-attention layers \cite{axial} outputting 512 channels; and finally a classification head outputting 12 channels, one for each precipitation rate bin. The total number of weights is 21M. The base model output logits are used as input for training the calibrator models.

{\bf Data.} The input data consists of 7 steps of MRMS radar images, 2 steps with 16 channels of GOES satellite, 1 step of precipitation prediction by HRRR (numerical weather model in North America); as well as topography, longintude, latitude, temporal information, and lead time. The target is MRMS discretized to 12 bins from 0.2 mm/h up to 10+ mm/h and one-hot encoded such that the base model predicts a 12-vector as a probability over the binned precipitation rates.

{\bf Calibration training data.} The data used in the development and evaluation of the calibration methods is temporally  non-overlapping from the data used for training of the base model. We use 110K unique samples (each sample has spatial size $64 \times 64$) to train the classifiers for flagging mispredictions in selective scaling. To optimize the temperatures in temperature scaling, selective scaling, and for learning the logit-temperature mapping in local temperature scaling, 1K samples are used. Finally, the uncalibrated and calibrated ETCE scores are computed over 47K samples not included in the training data of the calibrators.

\section{Results}

We summarize the ETCE scores averaged over the lead times in Table \ref{tab:avg-etce}. The first row corresponds to the base model. We include the number of learnable weights for the calibration methods; it includes classifier weights for the selective scaling schemes. The MLP-based selective scaling calibrator yielded the best improvement to model calibration with 23\% ETCE reduction. The Segformer-based selective scaling calibrator came as a close second with a 21\% reduction. Although temperature scaling has shown positive results in some computer vision problems, it did not reduce miscalibration here. Local temperature scaling was detrimental to model calibration, but the damage it did to calibration was alleviated when lead time encoding was used.

\begin{table}
    \centering
    \begin{tabular}{lcccc}
        \toprule
        \textbf{Calibrator} & \textbf{num. params} & \textbf{F1-score} & \textbf{avg. ETCE} & \textbf{$\Delta$ ETCE (\%)} \\
        \midrule
        Uncalibrated                     & $-$        & 0.565 & 0.079 & $-$   \\
        Temperature scaling              & 1          & 0.565 & 0.080 & -1.0  \\
        LTS (no lead time cond.)         & 2,107      & 0.573 & 0.096 & -21.3 \\
        LTS                              & 2,143      & 0.564 & 0.082 & -3.6  \\
        Selective scaling w/ MLP         & 3,254      & 0.564 & \textbf{0.060} & \textbf{23.5} \\
        Selective scaling w/ Segformer B0& 3,728,550  & 0.567 & 0.062 & 21.6  \\
        \bottomrule
    \end{tabular}
    \caption{ETCE scores averaged over all lead times for different calibrator models and the relative improvement over the uncalibrated baseline model. We also show the average F1-score computed for thresholded precipitation at 1 mm/h.}
    \label{tab:avg-etce}
\end{table}

\begin{figure}
    \centering
    \includegraphics[width=0.7\linewidth]{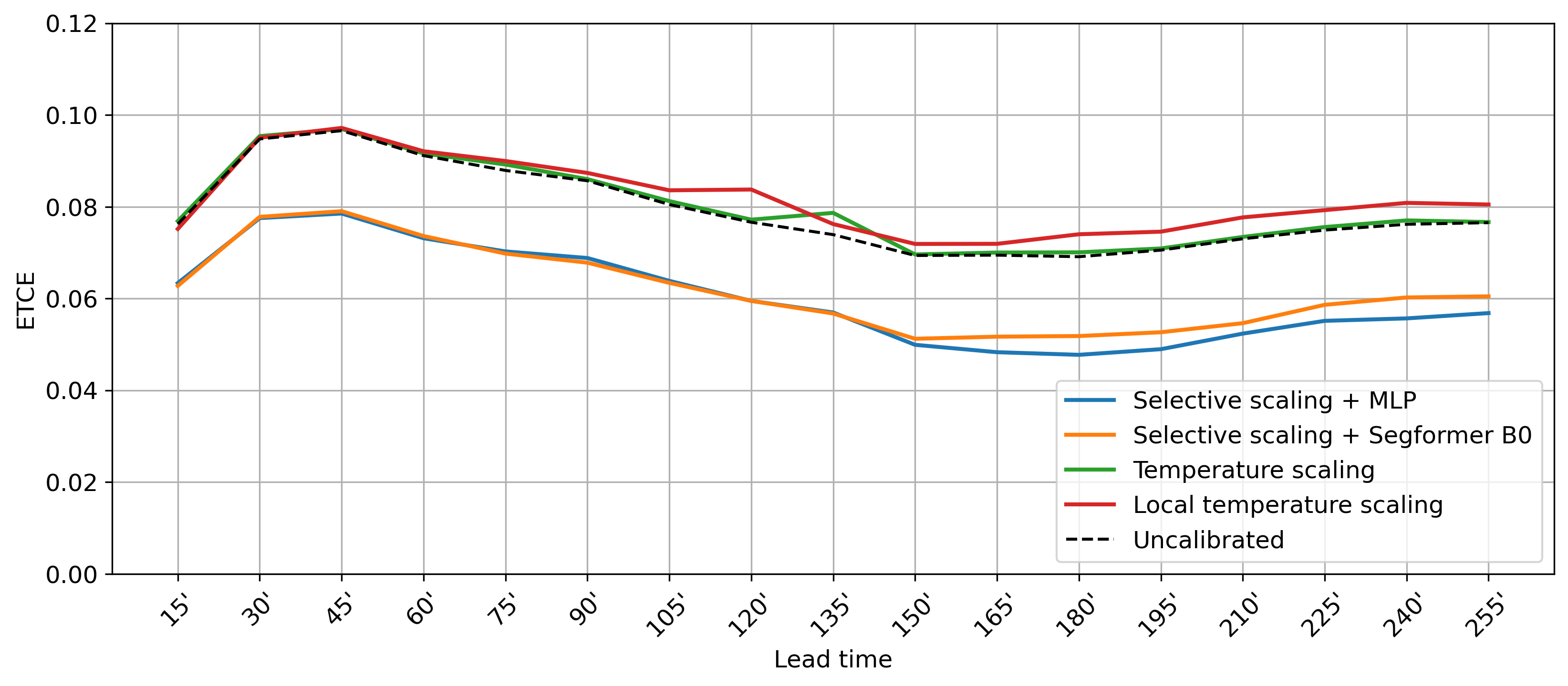}
    \caption{ETCE as a function of lead time for the uncalibrated model, and after applying temperature scaling, local temperature scaling and selective scaling.}
    \label{fig:results}
\end{figure}

A more detailed ETCE comparison per lead time is shown in Figure \ref{fig:results}. Selective scaling reduced miscalibration effectively over all lead times. At shorter lead times up to 150 minutes, selective scaling with either MLP or Segformer-B0 classifiers reduced ETCE equally. But at longer lead times the MLP-based calibrator performed best. 
We also tested more complex classifiers with the selective scaling approach---Segformer-B1 and B2---and found only minor improvement in ETCE compared to MLP and B0 classifiers (See Figure \ref{fig:segformers} in the Appendix). The marginally increased accuracy with a Segformer-B2 classifier does not justify the massive increase in model complexity. A more detailed look on miscalibration reduction is shown in Figure \ref{fig:etce-change} in the Appendix.

\section{Conclusions}

We introduced the expected thresholded calibration error (ETCE) to measure probability calibration of probabilistic models in forecasting. Different from standard computer vision tasks, in forecasting there is the lead time dimension.  By combining selective scaling with lead time encoding we reduced the base model's calibration error by up to 23.5 \%. Based on these results, future works should build on selective scaling, possibly by conditioning the calibrator on spatial and/or temporal information in addition to lead time.

\clearpage
\bibliographystyle{unsrt} 
\bibliography{refs} 

\begin{thebibliography}{10}

\bibitem{gen-models}
Suman Ravuri, Karel Lenc, Matthew Willson, Dmitry Kangin, Remi Lam, Piotr
  Mirowski, Megan Fitzsimons, Maria Athanassiadou, Sheleem Kashem, Sam Madge,
  et~al.
\newblock Skilful precipitation nowcasting using deep generative models of
  radar.
\newblock {\em Nature}, 597(7878):672--677, 2021.

\bibitem{nowcast-challenge}
James~W. Wilson, Yerong Feng, Min Chen, and Rita~D. Roberts.
\newblock Nowcasting challenges during the beijing olympics: Successes,
  failures, and implications for future nowcasting systems.
\newblock {\em Weather and Forecasting}, 25(6):1691 -- 1714, 2010.

\bibitem{nowcastnet}
Y.~Zhang, M.~Long, and K.~et~al Chen.
\newblock Skilful nowcasting of extreme precipitation with nowcastnet.
\newblock {\em Nature}, 619:526--532, 2023.

\bibitem{shi2025deeplearningfoundationmodels}
Jimeng Shi, Azam Shirali, Bowen Jin, Sizhe Zhou, Wei Hu, Rahuul Rangaraj,
  Shaowen Wang, Jiawei Han, Zhaonan Wang, Upmanu Lall, Yanzhao Wu, Leonardo
  Bobadilla, and Giri Narasimhan.
\newblock Deep learning and foundation models for weather prediction: A survey,
  2025.
\newblock arXiv preprint arXiv:2501.06907,
  \url{https://arxiv.org/abs/2501.06907}.

\bibitem{aabonev2023modelling}
Boris Bonev, Thorsten Kurth, Christian Hundt, Jaideep Pathak, Maximilian Baust,
  Karthik Kashinath, and Anima Anandkumar.
\newblock Modelling atmospheric dynamics with spherical fourier neural
  operators.
\newblock In {\em ICLR 2023 Workshop on Tackling Climate Change with Machine
  Learning}, 2023.

\bibitem{mcnally2024datadrivenweatherforecasts}
Anthony McNally, Christian Lessig, Peter Lean, Eulalie Boucher, Mihai Alexe,
  Ewan Pinnington, Matthew Chantry, Simon Lang, Chris Burrows, Marcin Chrust,
  Florian Pinault, Ethel Villeneuve, Niels Bormann, and Sean Healy.
\newblock Data driven weather forecasts trained and initialised directly from
  observations, 2024.
\newblock arXiv preprint arXiv:2407.15586,
  \url{https://arxiv.org/abs/2407.15586}.

\bibitem{bi2023accurate}
Kaifeng Bi, Lingxi Xie, Huan Zhang, et~al.
\newblock Accurate medium-range global weather forecasting with 3d neural
  networks.
\newblock {\em Nature}, 619:533--538, 2023.

\bibitem{andrychowicz2023deeplearningdayforecasts}
Marcin Andrychowicz, Lasse Espeholt, Di~Li, Samier Merchant, Alexander Merose,
  Fred Zyda, Shreya Agrawal, and Nal Kalchbrenner.
\newblock Deep learning for day forecasts from sparse observations, 2023.
\newblock arXiv preprint arXiv:2306.06079,
  \url{https://arxiv.org/abs/2306.06079}.

\bibitem{lang2024aifsecmwfsdatadriven}
Simon Lang, Mihai Alexe, Matthew Chantry, Jesper Dramsch, Florian Pinault,
  Baudouin Raoult, Mariana C.~A. Clare, Christian Lessig, Michael Maier-Gerber,
  Linus Magnusson, Zied~Ben Bouall{\`e}gue, Ana~Prieto Nemesio, Peter~D.
  Dueben, Andrew Brown, Florian Pappenberger, and Florence Rabier.
\newblock Aifs -- ecmwf's data-driven forecasting system, 2024.
\newblock arXiv preprint arXiv:2406.01465,
  \url{https://arxiv.org/abs/2406.01465}.

\bibitem{adrian2025data}
Melissa Adrian, Daniel Sanz-Alonso, and Rebecca Willett.
\newblock Data assimilation with machine learning surrogate models: A case
  study with fourcastnet.
\newblock {\em Artificial Intelligence for the Earth Systems}, 4(3):e240050,
  2025.

\bibitem{price2025probabilistic}
Iain Price, Alvaro Sanchez-Gonzalez, Ferran Alet, et~al.
\newblock Probabilistic weather forecasting with machine learning.
\newblock {\em Nature}, 637:84--90, 2025.

\bibitem{Nixon_2019_CVPR_Workshops}
Jeremy Nixon, Michael~W. Dusenberry, Linchuan Zhang, Ghassen Jerfel, and Dustin
  Tran.
\newblock Measuring calibration in deep learning.
\newblock In {\em Proceedings of the IEEE/CVF Conference on Computer Vision and
  Pattern Recognition (CVPR) Workshops}, June 2019.

\bibitem{guo}
Chuan Guo, Geoff Pleiss, Yu~Sun, and Kilian~Q. Weinberger.
\newblock On calibration of modern neural networks.
\newblock In {\em Proceedings of the 34th International Conference on Machine
  Learning - Volume 70}, ICML'17, page 1321–1330. JMLR.org, 2017.

\bibitem{wang10205155}
Dongdong Wang, Boqing Gong, and Liqiang Wang.
\newblock { On Calibrating Semantic Segmentation Models: Analyses and An
  Algorithm }.
\newblock In {\em 2023 IEEE/CVF Conference on Computer Vision and Pattern
  Recognition (CVPR)}, pages 23652--23662, Los Alamitos, CA, USA, June 2023.
  IEEE Computer Society.

\bibitem{NEURIPS2019_f8c0c968}
Ananya Kumar, Percy~S Liang, and Tengyu Ma.
\newblock Verified uncertainty calibration.
\newblock In H.~Wallach, H.~Larochelle, A.~Beygelzimer, F.~d\textquotesingle
  Alch\'{e}-Buc, E.~Fox, and R.~Garnett, editors, {\em Advances in Neural
  Information Processing Systems}, volume~32. Curran Associates, Inc., 2019.

\bibitem{gawlikowski2023survey}
Jakub Gawlikowski, Christopher R.~N. Tassi, Muhammad Ali, et~al.
\newblock A survey of uncertainty in deep neural networks.
\newblock {\em Artificial Intelligence Review}, 56(Suppl 1):1513--1589, 2023.

\bibitem{9710154}
Zhipeng Ding, Xu~Han, Peirong Liu, and Marc Niethammer.
\newblock Local temperature scaling for probability calibration.
\newblock In {\em 2021 IEEE/CVF International Conference on Computer Vision
  (ICCV)}, pages 6869--6879, 2021.

\bibitem{ding_local}
Zhipeng Ding, Xu~Han, Peirong Liu, and Marc Niethammer.
\newblock Local temperature scaling for probability calibration.
\newblock In {\em 2021 IEEE/CVF International Conference on Computer Vision
  (ICCV)}, pages 6869--6879, 2021.

\bibitem{hinton2015distillingknowledgeneuralnetwork}
Geoffrey Hinton, Oriol Vinyals, and Jeff Dean.
\newblock Distilling the knowledge in a neural network, 2015.
\newblock arXiv preprint arXiv:1503.02531,
  \url{https://arxiv.org/abs/1503.02531}.

\bibitem{balanya2024adaptive}
Sergi~A. Balanya, Javier Maro{\~n}as, and Daniel Ramos.
\newblock Adaptive temperature scaling for robust calibration of deep neural
  networks.
\newblock {\em Neural Computing and Applications}, 36:8073--8095, 2024.

\bibitem{perez2017filmvisualreasoninggeneral}
Ethan Perez, Florian Strub, Harm de~Vries, Vincent Dumoulin, and Aaron
  Courville.
\newblock Film: Visual reasoning with a general conditioning layer, 2017.
\newblock arXiv preprint arXiv:1709.07871,
  \url{https://arxiv.org/abs/1709.07871}.

\bibitem{axial}
Jonathan Ho, Nal Kalchbrenner, Dirk Weissenborn, and Tim Salimans.
\newblock Axial attention in multidimensional transformers.
\newblock {\em CoRR}, abs/1912.12180, 2019.

\end{thebibliography}

\clearpage
\appendix
\counterwithin{figure}{section}
\section{Appendix}

\subsection{ETCE details} \label{sec:details}

Here, we give the full details on the confidence and accuracy terms of Eq. \ref{eq:new-error}. In our dataset we have input-target pairs $(x_i, r_i)$ where $x_i$ is the DNN input and $r_i$ is the ground truth precipitation rate of sample $i$. Then, the term $\operatorname{conf}(b, R_k)$ for confidence bin $b$ and precipitation threshold $R_k$ is given by
\begin{align}
    \operatorname{conf}(b, R_k) \coloneq \frac{1}{\left| B_b \right|}\sum_{i\in B_b} s_{x_i}(R_k),\quad \text{where}\ B_b \coloneq \left\{i : s_{x_i}(R_k) \in b \right\},
\end{align}
where $s_{x_i}(\cdot)$ is the survivability function obtained from the predicted probability for input $x_i$. That is, $s_{x_i}(R_k) = 1-\operatorname{CDF}_{\operatorname{DNN}(x_i)}(R_k)$.

The accuracy term is
\begin{align}
    \operatorname{acc}(b, R_k) \coloneq \frac{1}{\left| B_b \right|} \sum_{i\in B_b} \mathbbm{1} (r_i \geq R_k),
\end{align}
where $r_i$ is the ground truth precipitation rate for sample $i$.

\subsection{Number of predictions in different confidence bins}
\begin{figure}[h!]
    \centering
    \includegraphics[width=0.75\linewidth]{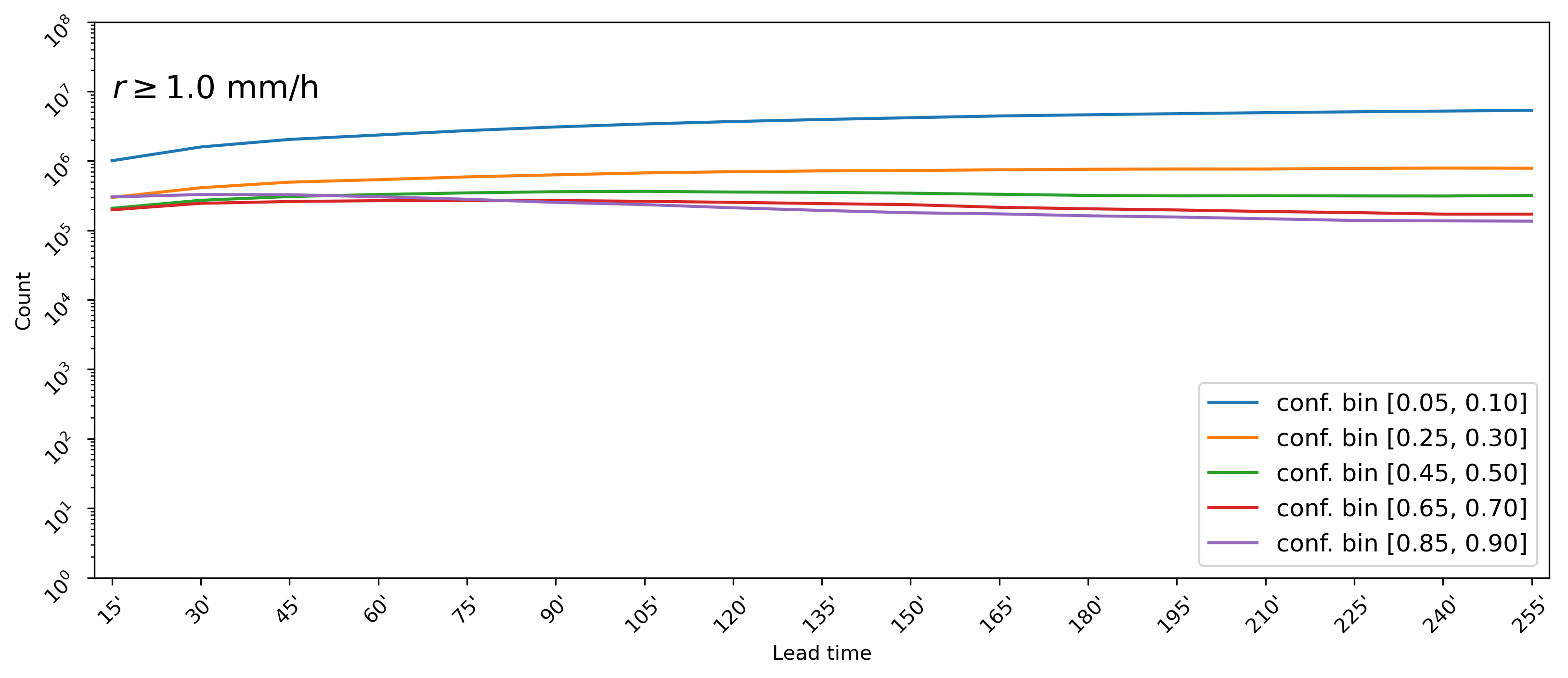}
    \caption{Number of predictions at threshold $r \geq 1.0$ mm/h within five selected confidence bins.}
    \label{fig:counts}
\end{figure}

Figure \ref{fig:counts} shows the number of predictions in five selected confidence bins. This example is for thresholded predictions at $r \geq 1.0$ mm/h.

\subsection{Selective scaling with Segformer classifiers}
\begin{figure}[h!]
    \centering
    \includegraphics[width=0.75\linewidth]{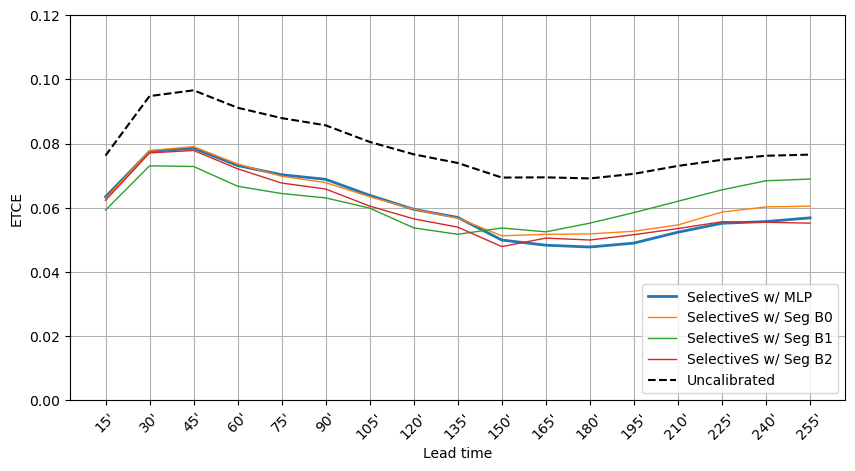}
    \caption{ETCE as a function of lead time for selective scaling using different classifiers for flagging mispredictions. Uncalibrated baseline shown with a dashed line.}
    \label{fig:segformers}
\end{figure}

Figure \ref{fig:segformers} shows ETCE as a function of lead time for selective scaling with different classifier models compared against the base model. Overall, using Segformer-B2 as the classifier resulted in approx. 1.3 \% lower ETCE compared to using the MLP but the significantly larger computational cost of the Segformer makes the MLP a more viable choice.

\subsection{Calibration effect on ETCE}
\begin{figure}[h!]
     \centering
     \begin{subfigure}[b]{0.95\textwidth}
         \centering
         \includegraphics[width=\textwidth]{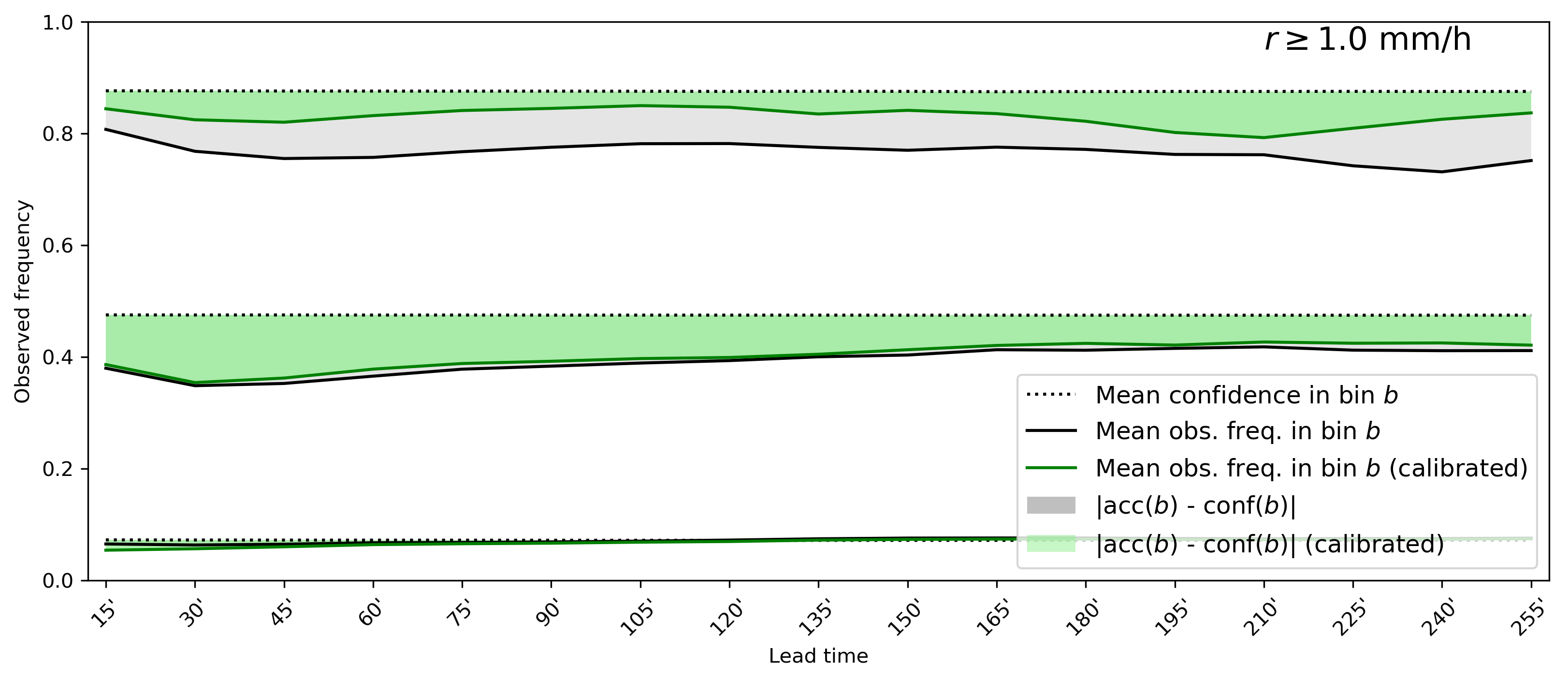}
     \end{subfigure}
     \\
     \begin{subfigure}[b]{0.95\textwidth}
         \centering
         \includegraphics[width=\textwidth]{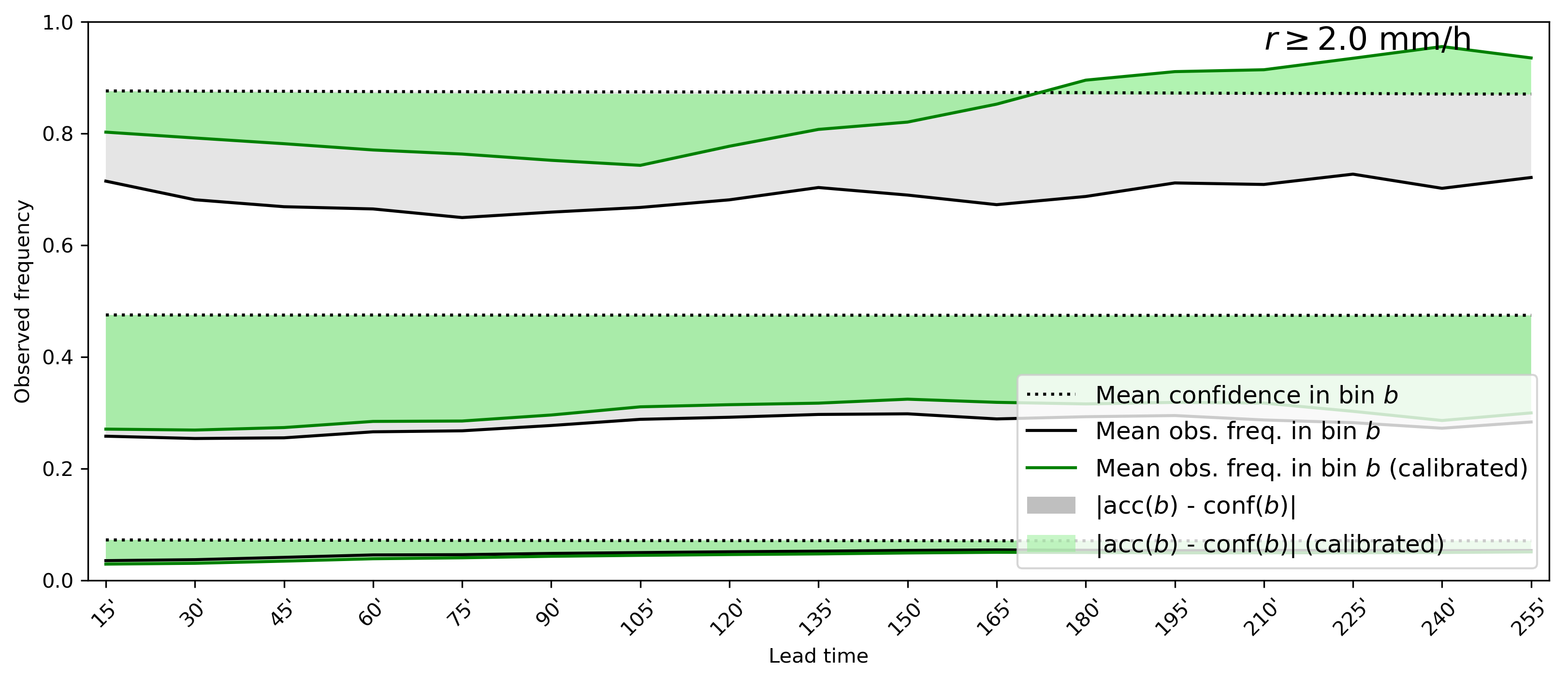}
     \end{subfigure}
        \caption{The difference between model confidence and accuracy for thresholds $r\geq1.0$ mm/h and $r \geq 2.0$ mm/h before and after applying calibration (Selective scaling with MLP classifier). Reduction in colored area shows reduction in miscalibration. Smallest area in between dashed and solid lines is best.}
        \label{fig:etce-change}
\end{figure}

Further detail on ETCE improvement under calibration is shown in Figure \ref{fig:etce-change}. Here, the specific calibration is selective scaling which we use with the MLP classifier for flagging mispredictions. The colored areas---gray for uncalibrated and green for calibrated---show the difference between average model confidence and accuracy at three thresholded confidence bins. For the uncalibrated baseline, we observe overconfidence across lead times and precipitation thresholds. The figure also shows that selective scaling reduces miscalibration effectively which shown by the smaller colored area. This is especially true when the predicted confidence is high.

\end{document}